# Automated Analytic Asymptotic Evaluation of the Marginal Likelihood for Latent Models

Dmitry Rusakov and Dan Geiger
Computer Science Department
Technion - Israel Institute of Technology
Haifa, Israel 32000
{rusakov,dang}@cs.technion.ac.il

## Abstract

We present two algorithms for analytic asymptotic evaluation of the marginal likelihood of data given a Bayesian network with hidden nodes. As shown by previous work, this evaluation is particularly hard because for these models asymptotic approximation of the marginal likelihood deviates from the standard BIC score. Our algorithms compute regular dimensionality drop for latent models and compute the non-standard approximation formulas for singular statistics for these models. The presented algorithms are implemented in Matlab and Maple and their usage is demonstrated on several examples.

## 1 INTRODUCTION

Asymptotic approximation of the marginal likelihood of data given a model is a critical part in many statistical applications. In particular, it is used in Bayesian model selection, where the model that maximizes the marginal likelihood of a given data is chosen (Cooper & Herskovits, 1992; Heckerman, Geiger & Chickering, 1995).

Recall that in the Bayesian approach to model selection, one chooses a model $M$ that maximizes the posterior model probability $P(M|D) \propto P(M,D) = P(M)P(D|M)$. Assuming prior model probabilities $P(M)$ are known, the problem of Bayesian model selection reduces to evaluating the marginal likelihood of the data $P(D|M)$ and maximizing it as a function of $M$. Since $P(D|M)$ is usually very small, it is more convenient to work on a logarithmic scale and optimize $\ln P(M,D) = \ln P(D|M) + \ln P(M)$. Furthermore, in practical settings $\ln P(D|M)$ is a large negative number that dominates the $\ln P(M)$ term for non-extremal $P(M)$ values. In other words, $\ln P(M,D) = \ln P(D|M) + O(1)$.

In this paper we address the problem of computing analytic asymptotic approximations of marginal likelihoods and present two computer programs that compute such approximations. Our algorithms are developed in the context of Bayesian networks with hidden variables, where the evaluation of marginal likelihood was shown to be particularly hard (Rusakov & Geiger, 2002).

Consider the evaluation of the marginal likelihood given a Bayesian network model. Under some regularity conditions, the asymptotic form of the log marginal likelihood for Bayesian network models without hidden variables is specified by the *standard BIC formula*:

$$\ln P(D|M) = \ln P(D|w_{ML}) - \frac{d}{2} \ln N + O(1), \quad (1)$$

where $N$ is the number of examples in $D$, $w_{ML}$ are the maximum likelihood parameters of $M$ and $d$ is the number of independent parameters. This asymptotic marginal likelihood formula was first developed by (Schwarz, 1978) for linear exponential models, which include undirected graphical models (Lauritzen, 1996). Later, this result was extended by (Haughton, 1988) for curved exponential models, which include the class of Bayesian network models without hidden variables (Geiger, Heckerman, King & Meek, 2001). Moreover, Haughton (1988) had showed that the standard BIC score is a consistent model selection criterion for curved exponential families.

The evaluation of the marginal likelihood for Bayesian network models with hidden variables is more complex. First, some of the parameters of the model may be redundant (Geiger, Heckerman & Meek, 1996), and second, some of the data may give rise to statistics that are *singular* relative to the given model, resulting in non-standard approximation forms (Rusakov & Geiger, 2002). Various heuristic methods were developed to approximate the marginal likelihood of Bayesian networks with hidden variables (Cheeseman & Stutz, 1995; Chickering & Heckerman, 1997).

In this work we address both difficulties. First, we implement the method for effective dimensionality computation presented in (Geiger, Heckerman & Meek, 1996) and optimize it by decomposing the input network into independent components. The algorithm is implemented in Matlab and is capable of evaluating effective dimensionality of large Bayesian networks with hidden variables.

Second, we fill in the details and implement the algorithmic approach suggested in (Watanabe, 2001) for analytic asymptotic approximation of "hard" integrals. Our algorithm combines state-of-the-art algorithms of algebraic geometry (Bodnar & Schicho, 2000; Bravo, Encinas & Villamayor, 2002) with specific analytic methods for marginal likelihood evaluation suitable for Bayesian networks, which were developed in (Rusakov & Geiger, 2002).

The latest algorithm is implemented in Maple and consists of a number of procedures. These procedures compute the approximation of marginal likelihood not only for Bayesian with hidden variables, but for a larger set of probabilistic models, for which the log-likelihood function can be represented (or bounded) by a polynomial. We demonstrate the usage of our algorithm in evaluating marginal likelihood formulas on a number of Bayesian networks with hidden variables and on other models.

The main contribution of this paper is in the connection and implementation of algorithmic ideas for analytic approximation of marginal likelihood integrals, thus paving the way for practical use of these techniques. The presented procedures are available at http://www.technion.ac.il/~rusakov/, along with commands that replicate all the results reported herein.

This paper is organized as follows. Sections 2 and 3 present background on Bayesian networks and asymptotic approximations of marginal likelihood integrals. Sections 4 and 5 present and evaluate algorithms for asymptotic marginal likelihood evaluation for regular and singular statistics of Bayesian networks with hidden variables. We conclude with a discussion in Section 6.

## 2 BAYESIAN NETWORK MODELS

Let $X = \{X_1, \ldots, X_n\}$ be $n$ discrete variables having $r_1, \ldots, r_n$ states respectively. A Bayesian network model $M$ for variables $X$ is a set of joint distributions for $X$ defined by a network structure $G_M$ and a set of local distributions $\mathcal{F}_M$. A probability distribution $P(x)$ belongs to a Bayesian network model $M$ if and only if it factors according to the network structure $G_M$ via

$$P(X = x) = \prod_{i=1}^{n} p_i(X_i = x_i | Pa(X_i) = j), \quad (2)$$

where $x$ is the $n$-dimensional vector of values of $X$, $Pa(X_i)$ denote parents of node $X_i$ in $G_M$, $j$ denote the values of $Pa(X_i)$ in $x$ and $p_i$ is a conditional distribution from $\mathcal{F}_M$. Intuitively, the graphical structure $G_M$ of model $M$ describes the dependencies (and independencies) of variables in distributions $P(X)$ in $M$.

We focus on discrete Bayesian networks, where local distributions $\mathcal{F}_M$ are multinomial distributions. We denote the *model* parameters defining $p_i(X_i = val_k | Pa(X_i) = j)$ by $w_{ijk}$ and the *joint space parameters* $P(X = x)$ by $\theta_x$. The mapping that relates these parameters, derived from Eq. 2, is

$$\theta_{(x_1, \ldots, x_n)} = \prod_{i=1}^{n} w_{ijk}, \quad (3)$$

where $k$ and $j$ denote the assignment to $X_i$ and $Pa(X_i)$ as dictated by $(x_1, \ldots, x_n)$. Given a Bayesian network model $M$, the number of model parameters is given by $ds = \prod_{i=1}^{n}(r_i - 1) \prod_{X_l \in Pa(X_i)} r_l$ and the number of joint space parameters is $dc = \prod_{i=1}^{n} r_i - 1$.

Consider now that in addition to the observable variables $X$, the model structure $G_M$ includes additional, unobserved variables $H = (H_1, \ldots, H_m)$. In this case, model $M$ consists of the distributions on $X$ that can be parameterized via

$$P(X = x) = \\ \sum_{H=h} \prod_{i=1}^{n} p_i(X_i = x_i | j) \prod_{l=1}^{m} p_l(H_l = h_l | j'),$$

where $j$ and $j'$ denote assignment given by $x$ and $h$ to the parents of $X_i$ and $H_l$. We denote the nodes of $G_M$ by $\tilde{X} = (\tilde{X}_1 = X_1, \ldots, \tilde{X}_n = X_n, \tilde{X}_{n+1} = H_1, \ldots, \tilde{X}_{n+m} = H_m)$ and we let $w_{ijk}$ be the model parameters of $M$ defined for all network nodes $\tilde{X}$. Consequently, the joint space parameters on the observable nodes $X$ are described by model parameters via the following formula:

$$\theta_{(x_1, \ldots, x_n)} = \sum_{(h_1, \ldots, h_m)} \prod_{i=1}^{n+m} w_{ijk}, \quad (4)$$

where $k$ and $j$ denote the assignment to $\tilde{X}_i$ and $Pa(\tilde{X}_i)$ as specified by assignments $(x_1, \ldots, x_n)$ to $X$ and $(h_1, \ldots, h_m)$ to $H$.

A typical example of Bayesian network with hidden node is the naive Bayesian network, where one hidden (class) node $H$ influences the observable (feature) nodes $X = (X_1, \ldots, X_n)$, as depicted on Figure 1. This network is used in clustering and classification (Cheeseman & Stutz, 1995).

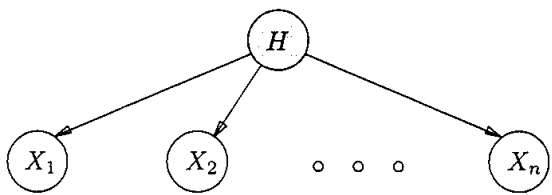

Figure 1: A naive Bayesian model. Node $H$ is hidden.

## 3　REVIEW OF ASYMPTOTIC APPROXIMATION OF MARGINAL LIKELIHOOD

Consider the marginal likelihood of data $D$ given a Bayesian network model $M$. The marginal likelihood $P(D|M)$ can be written as a function of the averaged sufficient statistics $Y_D = (Y_1, \ldots, Y_{dc})$ (i.e., normalized counts) and the number $N$ of samples in $D$, namely,

$$P(D|M) = \mathbb{I}[N, Y_D] = \int_\Omega e^{-Nf(w,Y_D)} \mu(w) dw, \quad (5)$$

where $w$ are the model parameters from the domain $\Omega \subset [0,1]^{ds}$, $\mu(w)$ is the prior parameter density and $f(w, Y_D) = -loglikelihood(Y_D|w) = -\sum_x Y_x \theta_x(w)$ is minus the log-likelihood function of model $M$. Variables $\theta_x(w)$ are indexed by vectors in $r_1 \times \ldots \times r_n$ and are specified by Eqs. 3 or 4 according to the type of the model.

When the model $M$ is a Bayesian model without hidden variables, $\theta_x(w)$ is defined by Eq. 3, and the prior parameter density $\mu(w)$ is usually assumed to be Dirichlet. Consequently, Integral 5 can be evaluated in closed form (Cooper & Herskovits, 1992; Heckerman, Geiger & Chickering, 1995).

When the prior parameter density is not known but supposed to be bounded, the approximation of $P(D|M)$ can be performed by bounding its value by two integrals using upper and lower bounds on $\mu$. The two bounding integrals can then be evaluated resulting in $O(1)$ approximation to $\ln P(D|M)$, as specified by Eq. 1.

The evaluation of Integral 5 for $M$ being a Bayesian network model with hidden variables is more difficult, because it is complicated by two factors. First, some of the model parameters may be redundant yielding a surface of maximum likelihood points in the model parameter space $\Omega$, rather than a single point. This can be tested by considering the Jacobian of Transformation 4, whose regular rank is the number of independent parameters of the model (Geiger, Heckerman & Meek, 1996). Alternatively, this case can be checked directly by considering the set of solutions of polynomial equations for maximum likelihood model parameters. We call statistics that yield maximum likelihood points surfaces without self-crossings, regular statistics. Note that when the model has no redundant parameters all statistics are regular and there are only a finite number maximum likelihood points. (It is not a single point, due to symmetry in the states of the hidden nodes.)

In the neighborhood of each maximum likelihood point on the regular maximum likelihood surface the log-likelihood function can be approximated by a quadratic form on the independent variables. Consequently, using the classical Laplace approximation procedure, e.g., (Wong, 1989), one obtains that $\ln P(D|M)$ is approximated by the standard BIC score, Eq. 1, with $d$ being the number of independent parameters of $M$. The number of independent parameters of the model, called the effective dimensionality, is computed by evaluating the regular rank of Jacobian of Transformation 4 (Geiger, Heckerman & Meek, 1996). We present an optimized procedure for evaluating the effective dimensionality of an arbitrary Bayesian model $M$ in Section 4.

Second, for some singular statistics, maximum likelihood surfaces intersect. Similarly, this can be tested by considering the Jacobian of Transformation 4, which is not of regular rank at the maximum likelihood surface intersection points (Geiger, Heckerman, King & Meek, 2001). In the neighborhoods of the intersection points the log-likelihood function can not be approximated by a quadratic form and the classical Laplace approximation of integrals fails. Recently, an advanced method for evaluating this type of integrals was introduced to the machine learning community by (Watanabe, 2001). It was shown that an approximation for $\ln P(D|M)$ is given by

$$\ln P(D|M) = \\ \ln P(D|w_{ML}) - \lambda \ln N + (m-1) \ln \ln N + O(1), \quad (6)$$

where $\lambda$ is the rational number less then or equal to half the number of independent parameters of $M$ and $m$ is an integer greater or equal to 1. A classical Laplace approximation to $\ln P(D|M)$ corresponds to $\lambda = \frac{d}{2}$ and $m = 1$.

The values of $\lambda$ and $m$ can be computed in a systematic way by computing the maximum pole and its multiplicity of the function

$$J(\lambda) = \int_{\Omega_\epsilon} [f(w, Y_D) - f(w_{ML}, Y_D)]^\lambda dw \quad (7)$$

evaluated in small neighborhoods $\Omega_\epsilon$ of maximum likelihood parameters $w_{ML} \in \Omega$. Computation of the integral $J(\lambda)$ can be performed by approximating the integrand function by a polynomial and changing $w$ to new coordinates in order to transform Integral 7 into a product of integrals of independent variables. The required change of coordinates is provided by the process of resolution of singularities from the field of algebraic geometry. Details are available in (Watanabe, 2001; Rusakov & Geiger, 2002).

The above method was applied to derive analytic asymptotic approximation of marginal likelihood for binary naive Bayesian model (Rusakov & Geiger, 2002). There it was shown that non-standard approximations to marginal likelihood, namely, approximations with $\lambda \neq \frac{d}{2}$ and $m \neq 1$, do happen for Bayesian networks with hidden variables.

Watanabe's method is based on combining the asymptotic theory with the machinery of resolution of singularities in algebraic geometry. Recent advances in computational algebraic geometry (Bodnar & Schicho, 2000; Bravo, Encinas & Villamayor, 2002) made possible to implement this method. Section 5 explicates the algorithmic details of the implementation that automatically produces analytic approximation of "hard" marginal likelihood integrals, using methods and procedures of algebraic geometry.

## 4 AUTOMATIC EFFECTIVE DIMENSIONALITY COMPUTATIONS

In this section we describe our algorithm to compute the effective dimensionality of a given Bayesian network and evaluate its performance. The algorithm, shown on Figure 2, is based on the decomposition of a given Bayesian network into independent components and evaluation of the Jacobian rank of Transformation 4 for each sub-network. The actual Jacobian rank need only be evaluated for the Markov neighborhoods of each of the hidden nodes, thus preventing the construction of prohibitively large matrices as would be required by the standard approach (Geiger, Heckerman & Meek, 1996). This idea of decomposing the network has been influenced by the idea of model decomposition for models with isolated hidden nodes (Settimi & Smith, 1999).

Our algorithm is implemented in Matlab and it is available as a Matlab package from http://www.technion.ac.il/~rusakov/bndim/.
The input to the algorithm is a Bayesian network structure, namely, the set of vertices and edges of a Bayesian network, and it does not require any manual coding of derivatives, in contrast to early computer aided computations of effective dimensionality (Geiger, Heckerman & Meek, 1996).

The implemented procedure is useful for evaluating

**Input:**
   $G_M$ - Bayesian network structure.
   $r = (r_1, \ldots, r_n)$ - Number of states of each node.
**Algorithm:**
1. Find disjoint Markovian neighborhoods of the hidden nodes.
2. For each neighborhood $G_i$ do
   2.1 Set $ds_i = \prod_{\tilde{X}_j \in G_i}(r_j - 1) \prod_{\tilde{X}_k \in Pa_{G_i}(\tilde{X}_j)} r_k$.
   2.2 Construct Jacobian $\mathcal{J}$ of Transformation 4.
   2.3 Evaluate regular rank of $\mathcal{J}$ by substitution of 3 random values into $\mathcal{J}$ and evaluation of numerical rank.
   2.4 Effective dimensionality of $G_i$, $de_i$, is with probability 1 the maximal numerical rank of $\mathcal{J}$ computed in previous stage at 3 random points.
   2.5 Set $\Delta d_i = ds_i - de_i$.
3. Set $ds = \prod_{i=1}^{n}(r_i - 1) \prod_{\tilde{X}_j \in Pa(\tilde{X}_i)} r_j$.
4. Return: effective dimensionality is $ds - \sum_i \Delta d_i$.

Figure 2: Effective dimensionality computation algorithm.

effective dimensionality of various Bayesian networks with hidden variables and as a research tool for testing closed form formulas for effective dimensionality of different types of latent models.

### 4.1 EVALUATION: NAIVE BAYESIAN MODELS

We explored the space of naive Bayesian networks to find networks that have effective dimensionality less than the number of model parameters and less than the number of joint space parameters. We call such models *degenerate*. We have chosen naive Bayesian networks as a particular test case for our procedure, since these models, being the most simple of the latent models, have been investigated to some extent by previous works. In particular, we evaluate for each model an upper bound on the effective dimensionality derived by (Kocka & Zhang, 2002) using the results of (Settimi & Smith, 1998). We call this bound the KZ-bound and study its tightness on degenerate models that are found. The results suggest several conjectures regarding the tightness of the KZ-bound and the effective dimensionality of naive Bayesian models.

The space of naive Bayesian models is parameterized by the number of classes, number of features and number of states of each feature. First, we explore naive Bayesian networks with three feature nodes. The results of search for degenerate models among the models with less than or equal to 7 states in all variables are summarized in the upper part of Table 1. The notation (3:2,2,4) in this table denotes the naive Bayesian model with 3 hidden states and 3 feature nodes with 2, 2 and 4 states respectively. The search on these 336 different models took about 25 minutes on Matlab 6 R13, Windows XP Pro, Pentium 1.6m, 512MB system. The average is 4.5 seconds per model and the actual time depends on the number of network and joint space parameters of the model, see Table 1. Another search, on all three-features naive Bayesian models with less than 10 states, took 17 hours and resulted in finding of 49 degenerate models in addition to 17 found by the previous search. Of these models, only one model has non-exact KZ-bound, as shown at the bottom of Table 1. Third, we explored naive Bayesian network with four feature nodes and less than 6 hidden states. The degenerate models found are also shown at the bottom of Table 1.

Table 1: Some of the degenerate naive Bayesian models found by the effective dimensionality algorithm. Legend: de - effective dimensionality, $ds$ - number of model parameters, $dc$ - number of joint space parameters, kz - KZ-bound. The model is described by giving the number of hidden states followed by the number of states of the feature nodes. Dots indicate that all models with the last feature variable having more states than displayed are also degenerate. Models for which the KZ-bound is not tight are in bold.

| NB model | de | $ds$ | $dc$ | kz | Time(sec) |
|---|---|---|---|---|---|
| 3:2,2,4... | 14 | 17 | 15 | 14 | 0.45 |
| 4:2,3,5... | 27 | 31 | 29 | 27 | 1.26 |
| 5:2,3,6... | 34 | 44 | 35 | 34 | 1.17 |
| 5:2,4,6... | 44 | 49 | 47 | 44 | 1.59 |
| 6:2,4,7 | 53 | 65 | 55 | 53 | 2.53 |
| 6:2,5,7 | 65 | 71 | 69 | 65 | 3.33 |
| 6:3,3,7 | 59 | 65 | 62 | 59 | 2.92 |
| **4:3,3,3** | **25** | 27 | 26 | 26 | 1.23 |
| **5:3,4,4** | **43** | 44 | 47 | 47 | 2.82 |
| **7:3,5,5** | **73** | 76 | 74 | 74 | 4.64 |
| **10:3,7,7** | **145** | 149 | 146 | 146 | 27.69 |
| 3:2,2,2,2 | 13 | 14 | 15 | 14 | 0.65 |
| 5:2,2,3,3 | 33 | 34 | 35 | 34 | 2.92 |
| 6:2,2,2,7... | 53 | 59 | 55 | 53 | 6.09 |

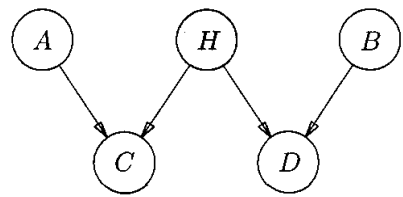

Figure 3. W-structure. Node $H$ is hidden.

The results of this search suggests a number of conjectures on the behavior of effective dimensionality of naive Bayesian models and on the tightness of the KZ-bound. In particular, we observe that the model is always degenerate when it has more parameters than the number of joint space parameters and one of the feature nodes have more states than the hidden node. This is intuitive because in such cases the hidden node does not have enough states to pass the information from this special feature node to other feature nodes. Furthermore, in these cases, the effective dimensionality is correctly predicted by the KZ-bound. In contrast, in some cases, the model is degenerate even when all feature nodes have less states than the hidden nodes, and these cases are unpredictable by KZ-bound.

We have also explored naive Bayesian models with $n$ feature nodes having the same number of states each. We have searched networks with up to 10 binary features nodes and up to 20 hidden states, and for 5 node networks with up to 5 states in hidden and feature nodes. The results of these searches are negative; Except for the known (3:2,2,2,2) network, we found no degenerate naive Bayesian networks with the same number of states in the feature nodes. We are tempted to speculate that this might be true in general.

### 4.2 EVALUATION: OTHER MODELS

The applicability of our procedure is not limited to the naive Bayesian models. In this subsection we report effective dimensionality computations for a number of latent models considered by other authors.

First, we evaluate the effective dimensionality of the W-structure on binary nodes (Geiger, Heckerman & Meek, 1996), Figure 3. The evaluation of the effective dimensionality took 0.43 seconds and we confirm that the effective dimensionality of this model is 9, while the number of model parameters is 11 and the number of joint space parameters is 15. When the number of hidden states is 3 or more the effective dimensionality of the $W$-structure is 10 (tested for up to 100 hidden states).

Second, we considered the hierarchical latent class model with 5 binary feature nodes and 3 hidden nodes with 5, 3 and 3 states (Kocka & Zhang, 2002). The evaluation of the effective dimensionality took 40 seconds and we found that the effective dimensionality of this model is 23 while the number of model parameters is 41 and the number of joint space parameters is 31. This evaluation agrees with the theoretical result due to (Kocka & Zhang, 2002).

Finally, in order to demonstrate the efficiency of this approach we have evaluated the effective dimensionality of the ALARM network, when some of the nodes are made hidden. The ALARM network has been build by medical experts for monitoring patients in intensive care (Beinlich, Suermondt, Chavez & Cooper, 1989). This Bayesian network of 37 nodes connected by 46 edges consists of 509 model parameters and its joint parameter space contains $1.7 \cdot 10^{16}$ parameters. Disregarding the medical meaning of the nodes, we set nodes KINKEDTUBE and CATECHOL to be hidden and evaluated the effective dimensionality of the resulted network. It was evaluated to be 494 instead of 509 as the number of model parameters may suggest, and it took our program 18 seconds to compute this result.

## 5 AUTOMATIC MARGINAL LIKELIHOOD APPROXIMATION FOR SINGULAR STATISTICS

In this section we describe our algorithm for the automatic evaluation of the coefficients $\lambda$ and $m$ in the asymptotic approximation of the logarithm of marginal likelihood, as described in Section 3. The algorithm, presented on Figure 4, follows and generalizes the method used by (Rusakov & Geiger, 2002) for naive Bayesian networks and it uses the desing library for resolution of singularities developed by (Bodnar & Schicho, 2000).

The validity of the presented algorithm follows from Lemma 4 presented in (Rusakov & Geiger, 2002). This lemma allows us to simplify the log-likelihood function (step 6) and approximate it by a simple polynomial $p$ (step 7.1). This makes the subsequent resolution of singularities (step 7.2.1) possible and feasible. The validity of steps 7.2.1-3 is due to the results of (Watanabe, 2001).

This algorithm is implemented in the Maple computer system and available from the web-site http://www.technion.ac.il/~rusakov/resolution/.
The package includes also auxiliary procedures for determining the maximum likelihood parameters, detection of the singular/regular statistics type, and approximation of the general marginal likelihood integrals given by polynomial log-likelihood functions (not necessarily arising from Bayesian networks).

### 5.1 EVALUATION

Our algorithm was validated on a number of naive Bayesian models and on non Bayesian network models, where the log-likelihood function is a polynomial.

We have run the algorithm and found the asymptotic approximations for marginal likelihood for singular statistics given a naive Bayesian network, as specified in the upper part of Table 2. These results agree with the theoretical approximations as specified by Theo-

**Input**:
    $G_M$ - Bayesian network structure.
    $Y$ - averaged sufficient statistics of the sample.
**Algorithm**:
1. Find a single $w_{ML}$ - maximum likelihood model parameters given $Y$. Let $\theta_{ML} = \theta(w_{ML})$, Eq. 4.
2. Solve $\theta(w) = \theta_{ML}$ to get the set of all maximum likelihood parameters, $W_{ML} = \{w|\theta(w) = \theta_{ML}\}$.
3. Find the deepest singularity(ies) of $W_{ML}$, $W_0$.
4. If $W_{ML}$ has no singularities, return
$\lambda = -[ds - dim(W_{ML})]/2$ and $m = 1$.
5. If $W_0$ includes varieties of dimension greater than 0, choose a random representative point for each such variety.
6. Simplify $\theta$ as a function of $w$ by linear combination of existing $\theta$.
7. For each $w_0$ in $W_0$ do
   7.1 Consider polynomial $p = \sum[\theta(w) - \theta(w_0)]^2$.
   7.2 Evaluate asymptotic approximation coefficients $\lambda$ and $m$ of integral $I[N] = \int e^{-Np(w)}dw$:
      7.2.1 Resolve singularities at origin of variety defined by $p$ using desing package.
      7.2.2 Use the mappings created by desing to find poles of integral $J(\lambda) = \int p^\lambda dw$.
      7.2.3 Greatest pole $\lambda$ and its multiplicity $m$ give the coefficients of $\ln N$ and $\ln \ln N$ in approximation to $I[N]$, Eq. 6.
8. Return the greatest pair of coefficients $(\lambda, m)$ evaluated at step 7.

Figure 4: Algorithm for evaluation of the asymptotic approximation to marginal likelihood.

rem 3 of (Rusakov & Geiger, 2002), including the type 2 singularity of the two-node network, which has particularly non-standard form (namely, $\ln \ln N$ term in the approximation). The test run time is between a few seconds to a few minutes to compute the approximation parameters for each network, depending on the size of the network. The time in consecutive runs may be significantly smaller because our algorithm implements hashing of the computed resolution results.

However, on the larger networks, the time complexity of the resolution of singularities algorithm becomes prohibitively large. We are currently working on implementing additional optimizations and simplifications to allow the treatment of larger networks.

Steps 7.2.1-3 of algorithm 4 are implemented in a separate procedure which can be applied to evaluate the maximum likelihood of models with polynomial log-likelihood function. As an example of such computation we consider the asymptotic evaluation of the integral $I[N] = \int_{W_\epsilon} e^{-N \sum_{i \neq j} w_i^2 w_j^2} dw$ in $n$ variables $w_1, \ldots, w_n$. The evaluation of the asymptotic approximation of the logarithm of this integral (Table 2) agrees with the formula $\ln I[N] = -\frac{n}{4} \ln N + O(1)$ developed by (Rusakov & Geiger, 2002).

Table 2: Experimental results with Algorithm 4. Legend: "Marginal likelihood" - the marginal likelihood under evaluation. In the upper part of the table specified by number of nodes in the given naive Bayesian model and type of given statistics. In the lower part specified directly by log-likelihood $f$ function, Eq. 5. "$\lambda, m$" - asymptotic approximation parameters, Eq. 6.

| Marginal likelihood | $\lambda, m$ | Time |
|---|---|---|
| NB model, 1 feature node. | $-1/2, 1$ | 0.6 sec |
| NB, $n = 2$, $Y$ - sing. type 1. | $-3/2, 1$ | 0.8 sec |
| NB, $n = 2$, $Y$ - sing. type 2. | $-3/2, 3$ | 1.3 sec |
| NB, $n = 3$, $Y$ - regular. | $-7/2, 1$ | 3.5 sec |
| NB, $n = 3$, $Y$ - sing. type 1. | $-5/2, 1$ | 0.8 sec |
| NB, $n = 3$, $Y$ - sing. type 2. | $-2, 1$ | 6.1 sec |
| NB, $n = 4$, $Y$ - regular. | $-9/2, 1$ | 20 sec |
| NB, $n = 4$, $Y$ - sing. type 1. | $-7/2, 1$ | 20 sec |
| NB, $n = 4$, $Y$ - sing. type 2. | $-5/2, 1$ | 56 sec |
| NB, $n = 5$, $Y$ - regular. | $-11/2, 1$ | 4.5 min |
| NB, $n = 5$, $Y$ - sing. type 1. | $-9/2, 1$ | 4.6 min |
| NB, $n = 5$, $Y$ - sing. type 2. | — | > 2 hrs. |
| $f = \sum_{i \neq j; i,j=1}^{3} w_i^2 w_j^2$ | $-3/4, 1$ | 6 sec |
| $f = \sum_{i \neq j; i,j=1}^{4} w_i^2 w_j^2$ | $-1, 1$ | 26 sec |
| $f = \sum_{i \neq j; i,j=1}^{5} w_i^2 w_j^2$ | $-5/4, 1$ | 108 sec |
| $f = \sum_{i \neq j; i,j=1}^{6} w_i^2 w_j^2$ | $-3/2, 1$ | 26 min |

## 6  DISCUSSION

The contribution of this work is twofold. First, it generalizes, enhances and implements the effective dimensionality computations algorithm which is suitable for practical use. The usage of the algorithm is demonstrated and a number of previously unknown degenerate latent class models are reported. The current implementation is available on-line and used by several researchers.

Second, we have described and implemented an algorithm that computes analytic asymptotic approximations of marginal likelihood integrals that can not be approximated by the classical formulas. For this task we have established connections with algorithms from different fields of research. For now this algorithm is slow, due to the high time complexity of resolution of singularities procedure. Still, we were able to validate the algorithm on a number of simple latent class models. We view our algorithm and its implementation as an ongoing effort and we hope that the development of improved resolution of singularities algorithms will make our procedure more useful. The current algorithm and its implementation can serve as a research tool for those who are looking to develop closed form solutions for "hard" marginal likelihood integrals.

## Acknowledgements

We thank Gabor Bodnar, Thomas Kocka, Bernd Sturmfels and Sumio Watanabe for helpful discussions.